\documentclass[sigconf]{acmart}
\renewcommand\footnotetextcopyrightpermission[1]{} 
\usepackage{color}

\newcommand{\jm}[1]{\textcolor{blue}{jm: #1}}

\usepackage{amssymb}
\usepackage{amsmath}
\usepackage{verbatim}

\usepackage{algorithm, algorithmic}
\usepackage{subfig}

\usepackage{booktabs} 

\begin{document}
\title{Modeling and Simultaneously Removing Bias via Adversarial Neural Networks}

\author{John Moore*\textsuperscript{1}, Joel Pfeiffer\textsuperscript{1}, Kai Wei\textsuperscript{1}, Rishabh Iyer\textsuperscript{1}, Denis Charles\textsuperscript{1}, Ran Gilad-Bachrach\textsuperscript{1}}
\author{Levi Boyles\textsuperscript{1}, Eren Manavoglu\textsuperscript{1} \vspace{0.5em}}

\author{\{jomoor,joelpf\}@microsoft.com}
\author{\textsuperscript{1}Microsoft, Bellevue, WA}
\author{\small \textsuperscript{*}Work performed as a Microsoft Intern}
\renewcommand{\shortauthors}{Anonymous}

\begin{abstract}

In real world systems, the predictions of deployed Machine Learned models affect the training data available to build subsequent models.  This introduces a bias in the training data that needs to be addressed. Existing solutions to this problem attempt to resolve the problem by either casting this in the reinforcement learning framework or by quantifying the bias and re-weighting the loss functions. In this work, we develop a novel Adversarial Neural Network (ANN) model, an alternative approach which creates a representation of the data that is invariant to the bias. We take the Paid Search auction as our working example and ad display position features as the confounding features for this setting. We show the success of this approach empirically on both synthetic data as well as real world paid search auction data from a major search engine.

\end{abstract}



\maketitle

\section{Introduction}
A central task in an online advertising system is estimating the potential click-through rate (CTR) of an ad given a query, or \emph{PClick}.  Using this PClick estimate and an advertiser's bid, we run an auction to determine where we should place ads on a page.  These ad impressions and their corresponding features are used to train new PClick models (potentially in an online fashion \cite{trenches}).  Hence, online advertising suffers from a feedback loop where previously shown ads dominate the training set, and ads higher on the page comprise the majority of the positive samples (clicks).  This bias makes estimating a good PClick across all ads, queries and positions (or $\mathcal D$) difficult, due to the overrepresentation of features correlating with high click-through rate dominating the feature space.

	\begin{figure}[t]
    \centering
    \includegraphics[width=0.5\textwidth]{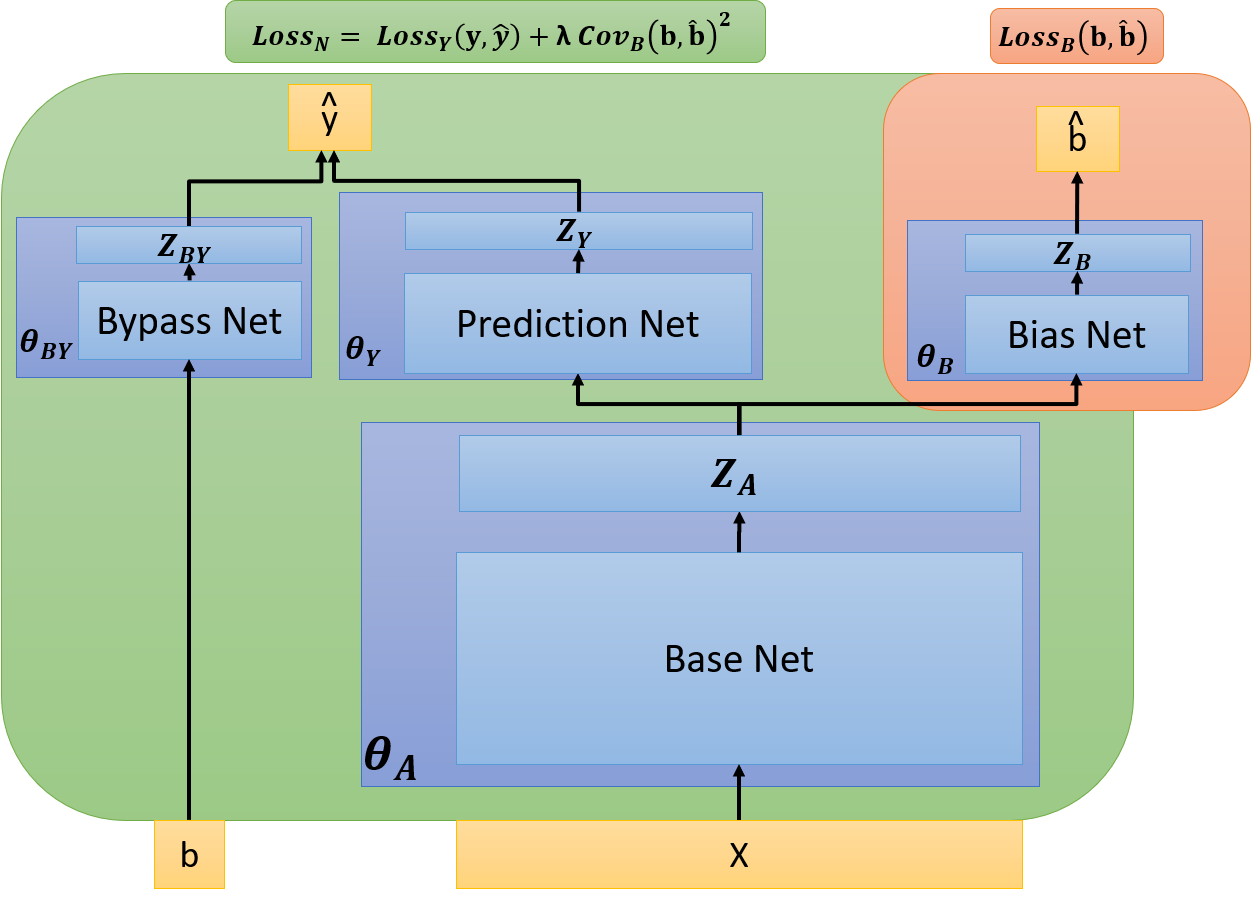}
    \caption{The Adversarial Neural Network representation best viewed in color.  The green area is optimized via $Loss_N$, which predicts the y variable (Click) and has a regularization for the distance of $b$ (position CTR) from noise.  Conversely, the orange parameters are optimized with respect to the bias network.  }
    \label{fig:main_network}
\end{figure}

We hypothesize that the position of an ad on a page (e.g., mainline, sidebar or bottom) can summarize a large portion of the PClick bias.  In effect, we aim to learn a PClick representation that is invariant to the position an ad is shown, that is, all potential ads retain a single relative ranking given a position on the page.  Although we can easily enforce this on the position feature itself by using a linear function, the intrinsic bias of the other features relative to position is not easily removed.

To learn this position invariant feature PClick model, we turn to {\em adversarial} neural networks (ANNs). ANNs are models with competing loss functions that are optimized in tandem (e.g., \cite{gans}), recent work \cite{advcrypt, pivot} has used them to hide or encrypt data.  Our ANN representation consists of four networks: Base, Prediction, Bias, and Bypass Networks (Figure \ref{fig:main_network}).  The final PClick prediction used online is the result of a linear combination of the outputs from the Bypass and Prediction networks to predict $\hat y$.  However, during training these predictors compete with the Bias network adversary.  This Bias network attempts to make predictions of the position using only the low rank vector $Z_A$ produced from the Base network.  Correspondingly, the Prediction, Bypass and Base networks optimize an augmented loss function that penalizes the Bias network.  The result is the vector $Z_A$ is largely uncorrelated with position before being passed into the Prediction network.

Other approaches to overcome position/display biases in online advertising exist, such as multi-armed bandit methods aid in generating less biased training data \cite{thompson, erenthompson} and covariate shift \cite{covariateshift}. However, each of these require sufficiently large samples from an exploration set to produce better estimates.  In practice, it is difficult to obtain sufficient amounts of exploration data as it typically impacts revenue significantly.  Our ANN approach requires no exploration and can be applied to an existing dataset.

To test the efficacy of the model, we show evaluations on real-world data and synthetic experiments.  We generate two sets of synthetic data to mimic the feedback loop present in an online Ads system, and show that systematic and user position biases are handled by the ANN to produce more accurate estimates.

We also demonstrate that there is a tradeoff between bias removal in the model while optimizing over CTR.  In evaluations, we show that by leveraging this tradeoff the ANN architecture has the ability to recover a more accurate estimate on unbiased datasets used in both synthetic and real-world datasets.

Our main contributions are the following:
\begin{itemize}  
\item A novel ANN representation for removing position bias in online advertising
\item Specifying a differentiable squared covariance loss to enable adversarial optimization over bias components.
\item Introducing a bypass structure to model position separately and linearly.
\item Detailed synthetic data generation evaluations to demonstrate the feedback problem present in online Ads systems.
\end{itemize}

\section{Position Bias in Paid Search}
\label{sec:background_1}

The feedback problem in ML applications is common. To demonstrate it, we focus on the problem of Click-through rate or PClick prediction in paid search advertising.
A standard Ad selection stack consists of a selection system, a model phase, and an auction phase\cite{eren}.  The selection system determines the subset of ads that are passed to the model.  The model  attempts to estimate the full probability density function across distribution $\mathcal{D}$, which is the entire Ads, Queries, and Positions space. Specifically, we estimate $P(Click|Ad, query, position)$.   In the auction phase, advertisers bid for keywords that are matched against queries.  Ads and their positions are finally selected given PClicks and advertiser bids.  We are mainly concerned about the model phase or PClick model in this work.  

It is difficult to estimate $\mathcal{D}$ for a couple of reasons. 
First, an online Ads system samples from a small, biased part of $\mathcal{D}$.  A machine learning model estimates PClick by using a variety of features across Ad and Query. Many of the rich features are counting features, which aggregate counting information across an Ad and Query's past (e.g. the percentage of clicks that this Ad/Query combination yielded in the past).  Query Ad pairs that are frequently presented in the Ads stack have rich informative feature information; however, Query Ad pairs that have never been seen or seen rarely will not have this rich information. Thus, it is naturally hard for a model to promote the ranking of a Query Ad pair that it has not shown online before, and the feedback loop continues.

Second, a feedback loop forms between training data and PClick model.  New training data or the ads that are subsequently shown online is formed from rankings from the previous model, and a new PClick model is formed from previous training data.   Thus, the resulting Feedback Loop (FL) data is biased towards past models and training data.

The Position Click-through rate, $P(y|Position=p)$, is the probability an ad is clicked given only the ad position on a page.  This is calculated by averaging the CTRs of ads shown online in a given position.  Ads in higher ranked positions typically yield higher CTRs.  Prior work has attempted to model or explain why position bias exists \cite{positionbias}. In our setting, we hypothesize that $P(y|Position=p)$ of past ads summarize much of these issues present in an online ads machine learning system since ads with higher Position CTRs are more likely to have an overrepresentation of features correlated with high PClicks.

In the ideal scenario, a PClick model will be trained only using a large amount of randomly and uniformly sampled (RUS) data from $\mathcal{D}$. A central goal of an online Ads stack, though, is ad revenue. In practice it is not possible to obtain a substantially large randomly sampled data set since it is costly to show many randomly paired Ads and queries online.

\section{Background}
\subsection{Online Advertising}
These issues with biased FL training data could be mitigated by framing the problem in terms of multi-armed bandits \cite{thompson}.  The central issue behind the multi-armed bandits problem is to find a reasonable Exploration and Exploitation tradeoff.  

In the Paid Search Advertising context, pulling an arm corresponds to selecting an ad to display \cite{erenthompson}.   Exploration practically means allowing ads with lower click probability estimates to sometimes appear online over ads with the highest estimates leading to a potential loss of short-term revenue.  Exploitation is preferring ads with the highest estimates typically resulting in immediate ad revenue gains.  

Bandit Algorithms have seen success in the display advertising literature and related areas such as news recommendation \cite{recommend, newsarticle}.  Thompson sampling is a popular method used in this literature that corresponds to drawing an arm according to its probability of being optimal and is preferred for its performance and simplicity \cite{thompson, thompsoneval, erenthompson}.  

These methods work best under the assumption that enough ads could be explored.  In an online machine learning system, this is increasingly not the case as medium-term and even short-term revenue losses are not acceptable.
A small sample of exploration data can be obtained, but it is generally too costly to obtain enough exploration data to have a substantial impact on the training set. Therefore, mostly biased FL data is still used for training a model, and these issues still remain.

Another approach to tackling this problem is answering the counterfactual question \cite{counterfactual}.  Bottou et al. show how to utilize counterfactual methodology from causal inference literature. Their methodology does not directly try to optimize performance on data sampled from $\mathcal{D}$, but it will rather estimate how different PClick models
would have performed in the past online.
The authors develop importance sampling techniques that estimate counterfactual expectations of interest with confidence interval bounds.  

Covariate shift is a related issue where the assumption is that $p(Y|X)$ remains the same across training and testing distributions where Y are labels and X are features. 
However, $p(X)$ shifts or changes from training to testing distributions.  Similar to counterfactual literature, there is work to rebalance the loss function in order to reflect this change in the test set by multiplying each instance by $w(x) = \frac{p_{test}(x)}{p_{train}(x)} $  \cite{covariateshift}. However, determining $w(x)$ whenever the test set does not have sufficient samples becomes difficult. The RUS dataset in our setting is not large enough to represent the entire distribution $\mathcal{D}$.


\subsection{Adversarial Networks}
Adversarial networks became popular recently, especially as a part of generative models in the context of Generative Adversarial Networks (GANs).  In GANs, the goal is to build a generative model that create realistic examples from some domain by optimizing two loss functions simultaneously between a generator and discriminator network \cite{gans}.  


Adversarial networks are used for other purposes as well. \cite{advcrypt} proposed using adversarial networks as a way to produce some level of encryption to data.  
The goal is to hide information from an adversary while being able to send information to a designated receiver.  
Neural Cryptographic systems do so by optimizing two loss functions in an adversarial fashion.  The first loss function can be seen as trying to encrypt the data, while the second attempts to decrypt the data adversarially.  The absolute covariance function can be defined as part of this encryption loss function.



In addition to encrypting data, adversarial optimization has been proposed when dealing with nuissance variables or variables that should not be correlated with the output prediction distribution \cite{pivot}.  This method uses a similar architecture and optimization technique as GANs.  However, instead of generating data, they penalize the first network if it produces predictions that can be used to predict the nuissance variable.  Similar to the discriminator, the second network attempts to predict the nuissance variable.  This work is distinct from ours for a couple of reasons. We are not interested in decorrelating predictions with position bias.  We are interested in a partial representation of features that are decorrelated with this bias, while still modeling  the bias. Furthermore, the training distribution derived from an online Ads stack is a biased sample from $\mathcal{D}$

\section{Method Description}

We develop an Adversarial Neural Network (ANN) architecture to produce accurate PClick predictions, $\mathbf{\hat{y}}$, given a biased Feedback Loop training set. We assume a continuous valued feature, $b$ that summarizes this bias. We define $b$ as position CTR or $P(y|Position=p)$ in the Ads context. A set of input features, $\mathbf X$ are typically weakly correlated with $b$.  



\subsection{Network Architecture}	
	The ANN representation consists of a Base, Prediction, Bias, and a Bypass Network shown in Figure \ref{fig:main_network} with parameters $\theta_{A}$, $\theta_Y$, $\theta_B$, $\theta_{BY}$ for each of the networks, respectively. 	
	 The first component, the Base and Prediction networks, is optimized to be $b$ independent, while the second component, the Bypass network, depends only on $b$.  By decomposing the model in this way, the ANN can learn from the data even when the bias exists.
	
	 The Bypass structure directly uses $b$ by incorporating its last hidden state $Z_{BY}$ as a linear term in the final sigmoid prediction function of equation \ref{eq:predp}. The set of final hidden states used for predicting $\hat{y}$ will consist of a linear combination of activations from both the Prediction and the Bypass Network. Let
	
\begin{equation} \label{eq:predp}
\hat{y} = sigmoid(W_Y Z_Y + W_{BY} Z_{BY} + c)
\end{equation}
    where $Z_Y$ refer the final hidden activations at the end of the Prediction network, $W_Y$ are the weights multiplied with $Z_Y$ and $W_{BY}$ is defined similarly for the Bypass Network.   $c$ is a standard linear offset bias term.
	
	This linear bypass strategy on $b$ allows the ANN to model $b$ separately and preserves the relative rankings across $b$'s (e.g. an ad will have a higher Click prediction if it has a higher $b$ value or position CTR) while directly incorporating $b$ 
	
	  Given $\mathbf X$, the Base Network outputs a set of hidden activations, $Z_A$ that are fed as inputs to both the Prediction and Bias networks as illustrated in Figure \ref{fig:main_network}. $Z_A$ is used to predict $y$ well, while predicting $b$ poorly.

\subsection{Loss Functions}
	To accomplish the desired set of hidden activations,   we minimize two competing loss functions, the bias loss, $Loss_B$, and the noisy loss, $Loss_N$.
	\begin{equation} \label{eq:lossp}
Loss_B(b, \hat{b}; \theta_B) = \sum_{i=0}^n (b_i - \hat{b}_i)^2
\end{equation}

	\begin{equation} \label{eq:lossn}
Loss_N(y, \hat{y}, b, \hat{b}; \theta_{A}, \theta_{BY}, \theta_Y) = (1-\lambda) Loss_Y(y, \hat y)  + \lambda \cdot Cov_B(b, \hat b)^2 
\end{equation}
	The bias loss function is defined in Equation \ref{eq:lossp}. This loss function measures how well the Bias network can predict $b$ given $Z_A$. 
	  In Figure \ref{fig:main_network}, only the Bias network (orange) and $\theta_B$ are optimized with respect to this loss function, while keeping all other parameters constant.  
	  
 Equation \ref{eq:lossn} describes the noisy loss function, which optimizes over $\theta_{A}$, $\theta_{BY}$, $\theta_Y$, while keeping $\theta_B$ constant.
	   This loss consists of $Loss_Y(y, \hat y)$ to represent the prediction loss and can be defined in various ways. In this work we define the $Loss_Y$ in terms of binary cross entropy.
	\begin{equation} \label{eq:lossc}
Loss_Y(y, \hat{y}) = \frac{1}{n} \sum_{i=0}^n y_i log(\hat{y}_i) + (1-y_i) log(\hat{y}_i)
\end{equation}   
	   
	   $Cov_B(b, \hat b)$ is a function of sample covariance and is computed by calculating means across $b$'s and $\hat{b}$'s in a given minibatch.
	\begin{equation} \label{eq:losscov}
Cov_B(b, \hat b)^2 = \left( \frac{1}{n-1} \sum_{i=0}^n (b_i - \bar{b}) (\hat{b}_i - \bar{\hat{b}})  \right)^2
\end{equation}

	   $Cov_B(b, \hat b)^2$ represents the distance $\hat b$ are from predicting noise.  
	  The squared covariance is 0 when $\hat b$ is not positively or negatively correlated with $b$.	  
	   When there is high correlation, $Loss_N$ would be highly penalized as long as $\lambda$ is sufficiently large.  
	
	The resulting $Loss_N$ objective function therefore penalizes the model for both poor predictions, and the ability for $\mathbf{X}$ to recover $b$ or (where an ad was placed on the page).  
 $\lambda$ controls how much each term is emphasized relative to the other. 

\subsection{Learning}
In practice, the covariance function is calculated across each minibatch individually where means are computed from each minibatch.
Both loss functions, $Loss_N$ and $Loss_B$ are optimized alternatively via stochastic gradient descent on the same minibatch (lines 5-6). 

\subsection{Online Inference}
To predict in an online setting or on a test set, we disregard the Bias network and use the other three networks to produce predictions, $\hat{y}$.  In the context of an online Ads system, we set $b$ to Position 1 CTR for data not seen online in the past, which is then fed into the Bypass network.

\begin{algorithm}[t]
\caption{Train($\mathbf{X}$, $\mathbf{C}$, $\mathbf{b}$, $maxItr$)}
{\scriptsize
\begin{algorithmic}[1]
\STATE Create Base, Prediction, Bias, and Bypass networks with $\theta_{A}$, $\theta_{Y}$, $\theta_B$, $\theta_{BY}$ corresponding to parameters of each network
\STATE Split $\mathbf{X}$, $\mathbf{Y}$, $\mathbf{B}$ into minibatches
\REPEAT
\STATE Optimize $Loss_N(y, \hat{y}, b, \hat{b}; \theta_{A}, \theta_{BY}, \theta_Y)$ with respect to $\theta_{A}$,$\theta_{BY}$, $\theta_Y$
\STATE Optimize $Loss_B(b, \hat{b}; \theta_B)$ with respect to $\theta_B$
\STATE $\mathbf{t_e}++$
  \UNTIL{$\mathbf{t_e} \geq maxItr$}
  

\RETURN DNN
\end{algorithmic}
}
\label{alg:rnnTra}
\end{algorithm}

\section{Synthetic Evaluations on System Level Bias}
\label{sec:synthetic1}

	\begin{figure}[t]
    \centering
    \includegraphics[width=0.5\textwidth]{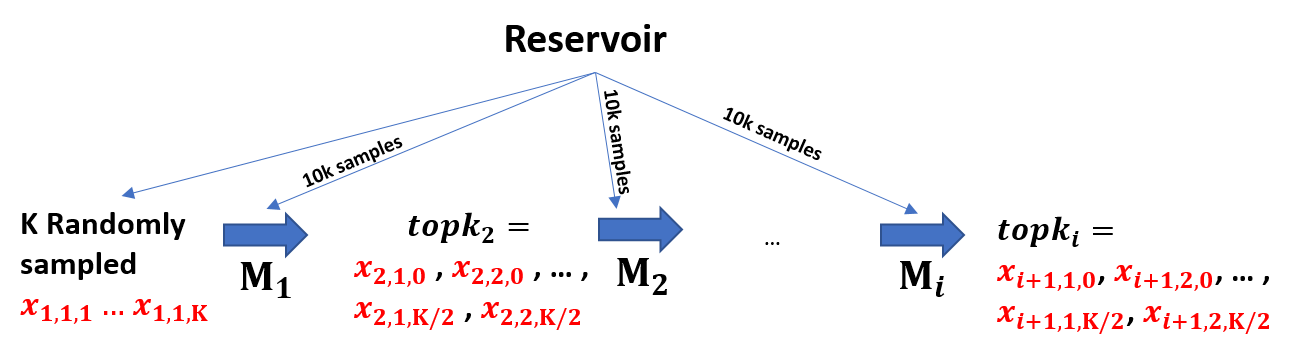}
    \caption{ Training data generated at each step of showing ads online.}
    \label{fig:feedback_loop}
\end{figure}

	\begin{figure}[t]
    \centering
    \includegraphics[width=0.3\textwidth]{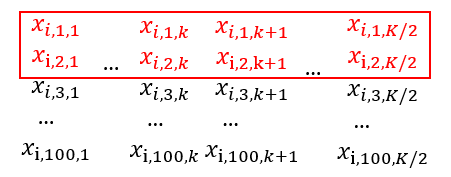}
    \caption{ Top 2 instances ranked by $M_{t-1}$ are selected from 10,000 candidates sampled without replacement}
    \label{fig:feedback_loop_2}
\end{figure}

\begin{table}
\centering
\begin{tabular}{c c}
Position 1 on $day_{T-1}$ & 0.464   \\
Position 2 on $day_{T-1}$ & 0.414 \\
Position 1 on $day_{T-2}$ & 0.454 \\
Position 2 on $day_{T-2 }$ & 0.396 \\
Position 1 over all days & 0.408 \\
Position 2 over all days & 0.378
\end{tabular}
\caption{\label{tab:CTRs} Position CTRs after System Level Bias synthetic evaluation}
\label{table:dataTable}
\end{table}

\begin{table}
\centering
\begin{tabular}{c c c}
Average Position CTR on FL (last 2 days) & 0.432 \\
MSE on FL using Average & 0.000782
\end{tabular}
\caption{\label{tab:method_MSE}  A naive approach that just predicts the Average CTR.  This forms an upper bound on MSE for FL data. }
\label{table:dataTable}
\end{table}

\begin{table}
\centering
\begin{tabular}{c c}
AUC & 0.775 \\
Log Loss & 0.277
\end{tabular}
\caption{\label{tab:method_comparison_AUC} Training and Testing using Logistic Regression on HeldOut data derived from $\mathcal{D}$. This forms an upper bound on AUC for the HeldOut set.}
\label{table:dataTable}
\end{table}

We generate synthetic data to illustrate the natural feedback loop or system level bias present in an online advertising stack.  We first generate click labels according to a bernoulli with probability $P(Y=1)=0.1$ where $Y=1$ represents a clicked ad.  Then, feature vectors, $\mathbf{x_j}$ are generated from two different but overlapping normal distributions according to

\[ \mathbf{x_j} =  \begin{cases} 
      N(0, \sigma) & \: if \: Y=0 \\
      N(1, \sigma) & \: if \: Y=1 
   \end{cases}
\]
where we set $\sigma=3$. 

 This process forms a complete distribution $\mathcal{D}$, and 100k samples are taken to form a large Reservoir dataset.
We then represent the feedback loop by simulating an iterative process where previously top ranked $\mathbf{x_j}$'s ( or ads) are used as training data to rank the next ads. Figures \ref{fig:feedback_loop} and \ref{fig:feedback_loop_2} shows this feedback loop process, and Algorithm \ref{alg:synthetic} demonstrates the simulation.  

$\frac{K}{2}$ candidate sets of 10,000 instances are drawn at random without replacement from the underlying Reservoir on day $i-1$ where K=500.
The model $M_{i-1}$ trained on day $i-1$ ranks the top 2 ads in each candidate set to show to the user on day $i$. 
Labels are revealed on day $i$, which subsequently forms the next iteration of available $topk_i$ training data.  

   We repeat this process until a desired number of iterations, T=100.  At each iteration, we record the average position CTR, $P(y|Position=p)$, for each of the top 2 positions.  $p=1$ refers to the top ranked ads, and $p=2$ are the 2nd top ranked ads.  We treat the position CTRs as the continuous bias term $b$. 
To start this process, we sample $K$ instances from the Reservoir to form $topk_0$ .  In an online Ads system, multiple days of training data are typically used to reduce systematic bias.  In the following evaluations we utilize the last two days of available training data (i.e. $M_i$ trains on $topk_i$ and $topk_{i-1}$).  Each model $M_i$ is a logistic regression classifier with l2 regularization.  
We set the parameter $r=0$ in line 13 of Algorithm \ref{alg:synthetic} to illustrate a system level feedback loop bias. 
We form testing data, separate from this feedback loop process, or HeldOut RUS evaluation by sampling 100k samples from $\mathcal{D}$.

CTRs for each position are shown in table \ref{tab:CTRs} on the last two days, and the overall CTRs calculated over all days.  All 4 CTR values are equally likely, since they are each associated with 250 training examples.  Therefore, a naive approach should predict the average CTR values. This forms an upper bound on how well an adversarial Bias Network can predict $b$.
 We record in table \ref{tab:method_MSE} the average CTR over the last two days of data (4 values) and calculate the MSE using this value.

\begin{algorithm}[t]
\caption{SyntheticFeedback($K$, $T$, $r$)}
{\scriptsize
\begin{algorithmic}[1]
\STATE Draw 100k labels according to $P(Y=1)=0.1$
\STATE Reservoir = 100k labels with 10 features $x_j$ according to
\STATE \[ \mathbf{x_i} =  \begin{cases} 
      N(0, \sigma) & \: if \: Y=0 \\
      N(1, \sigma) & \: if \: Y=1 
   \end{cases}
\]
\STATE HeldOut = draw a separate Reservoir with 100k samples
\STATE $topk_0$ = draw K samples from Reservoir and set $topk_{-1}=\emptyset$
\FOR{\texttt{(i = 0; i< T; i++)}}
\STATE Train $M_i$ on $topk_i$, $topk_{i-1}$
\STATE Set $topk_{i+1}$ to have 0 samples
\FOR{\texttt{(k = 0; k< K/2; k++)}}
	\STATE candidates = draw 10,000 samples from Reservoir
	\STATE Retrieve top 2 candidates (Position 1 and 2) using $M_i$
	\IF{Position 2 ad has Click==1}
		\STATE Set Position 2 ad Click=0 with probability $r$
	\ENDIF
	\STATE Add results to $topk_{i+1}$ 
\ENDFOR

\ENDFOR

\STATE Calculate $b_{T-1}$, $b_{T-2}$ or Position CTRs
\RETURN $topk_{T-1}$, $topk_{T-2}$, $b_{T-1}$, $b_{T-2}$, HeldOut
\end{algorithmic}
}
\label{alg:synthetic}
\end{algorithm}

\vspace{2cm}
\subsection{Setup}

We seek a model which is trained on FL data (i.e. the last two days of the synthetic generation process) but able to generalize to $\mathcal{D}$ or our RUS HeldOut data.   We train a set of ANNs using this FL data with different $\lambda$'s and set $b$ to its Position CTR.
The hyperbolic tangent function is used for all of the hidden activations except the last layers. The output activation function of the Prediction network is a sigmoid, and the output activation of the Bias Network is linear. The Bypass network consists of 1 hidden layer with 1 node, while the Base, Prediction, and Bias networks consist of 1 hidden layer with 10 nodes each.    We perform stochastic gradient descent with minibatch size=100 and a learning rate=0.01.  We train for 100 epochs (or passes) over the FL data.  After this main training process, we allow the Bias network to train over $Loss_B$ for 100 epochs.  Ideally, this allows the Bias network to do its best to predict $b$ given $Z_A$ produced from the Base network.

For comparison, we perform the same evaluations for an ANN with $\lambda=0$.  This model can be seen as a complete independent vanilla neural network optimizing over $y$, while a separate Bias network is able to observe and optimize $Loss_B$ without changing the Base Network.  We run 10 trials of each model with different weight initializations and report averaged Area under ROC curve (AUCs) on $y$ and averaged mean squared errors (MSEs) on $b$.  

\subsection{Main Synthetic Results}
\label{sec:synthetic_main_results}

\begin{figure*}[]
\captionsetup[sub]{font=small,labelfont={bf,sf}}
\centering
\subfloat[\footnotesize{AUC on the FL training set }]{\includegraphics[width=0.33\textwidth]{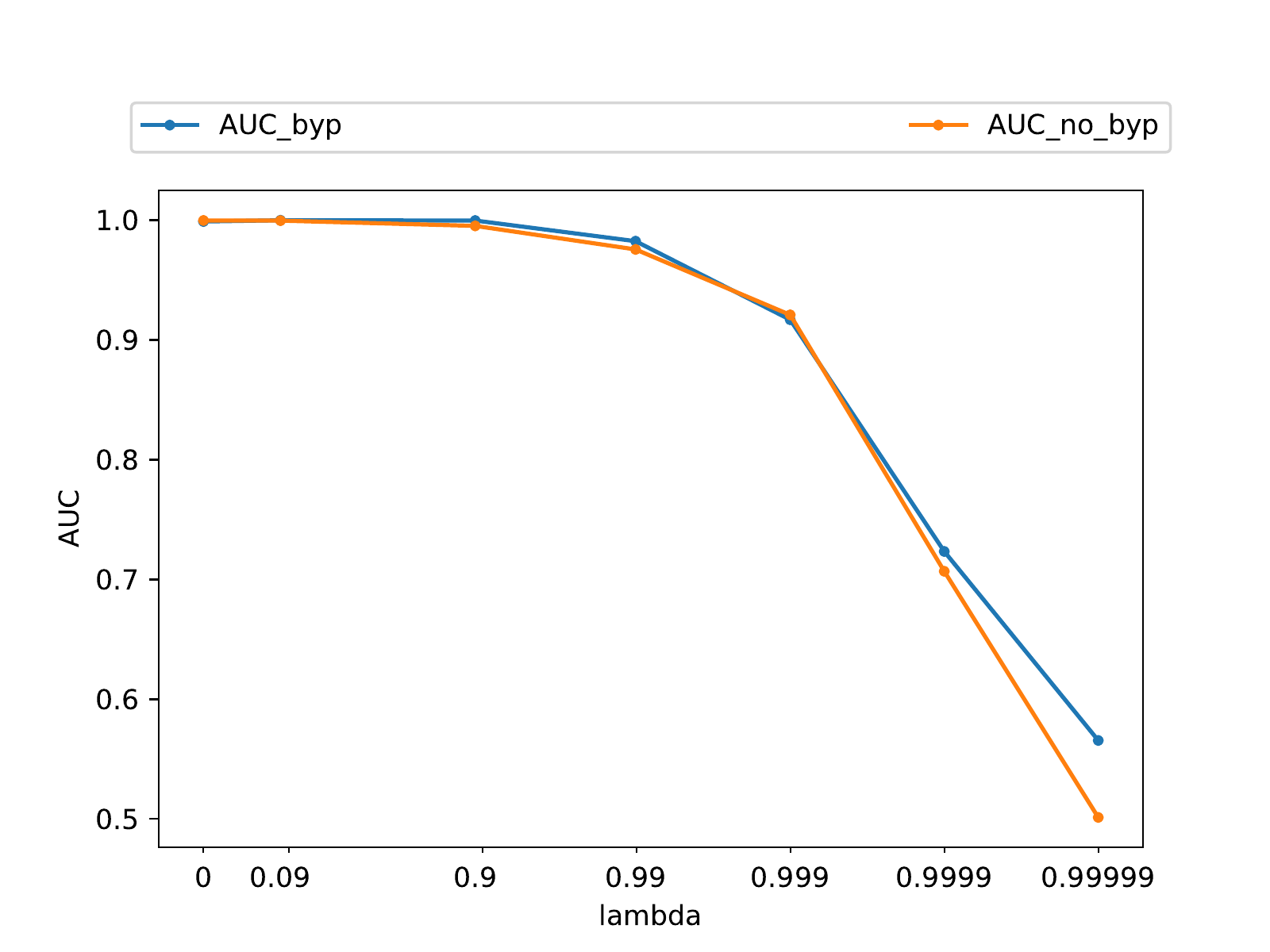}%
\label{fig:synthetic_auc_1}}
\subfloat[\footnotesize{MSE on the FL training set  }]{\includegraphics[width=0.33\textwidth]{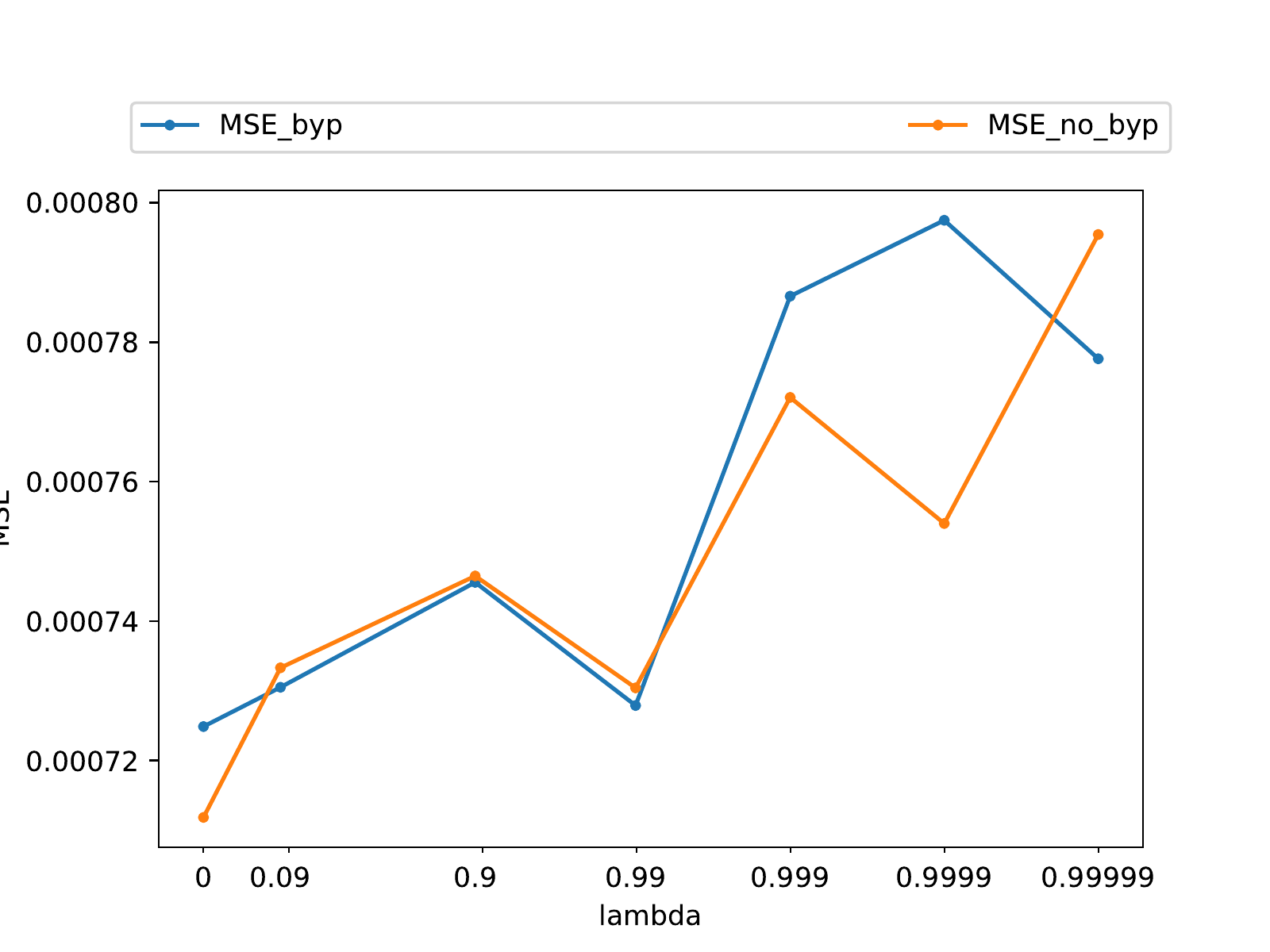}%
\label{fig:synthetic_mse_1}}
\subfloat[\footnotesize{AUC on the RUS testing set}]{\includegraphics[width=0.33\textwidth]{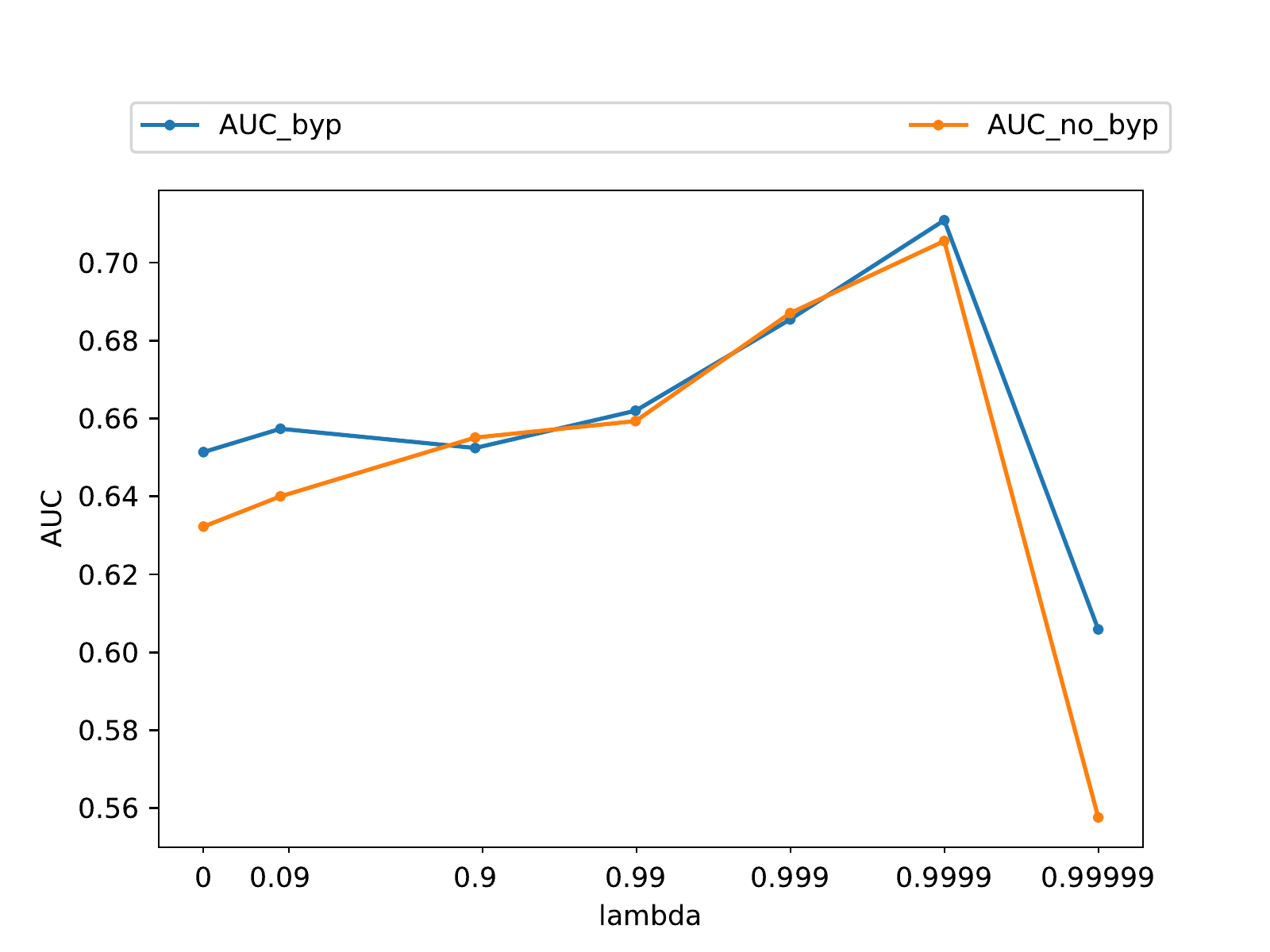}%
\label{fig:synthetic_auc_1}}

\subfloat[\footnotesize{MSE on the RUS testing set}]{\includegraphics[width=0.33\textwidth]{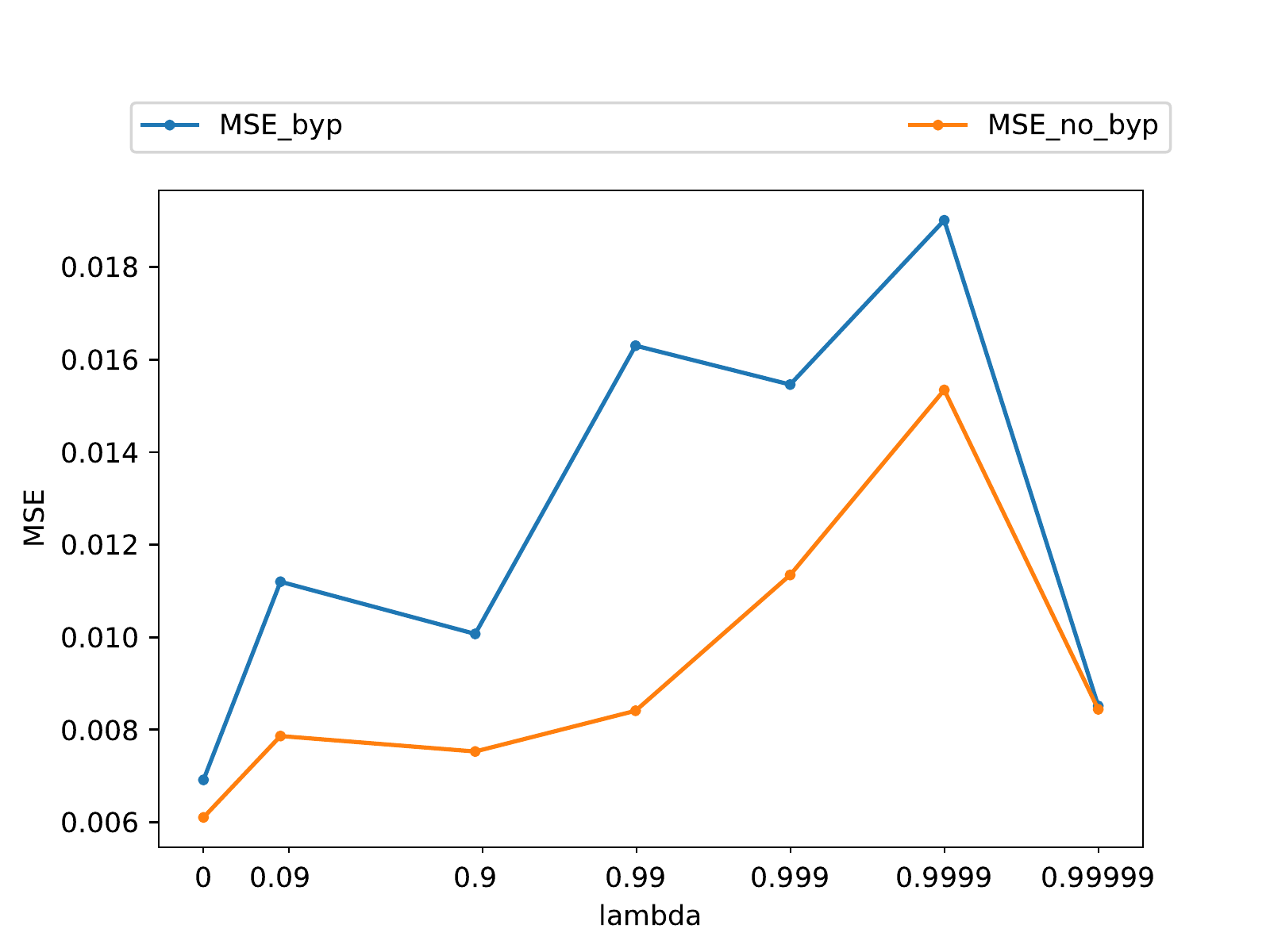}%
\label{fig:synthetic_mse_1}}
\subfloat[\footnotesize{Averaged differences in prediction between using Position 1 CTR vs Position 2 CTR in the linear bypass for RUS data}]{\includegraphics[width=0.33\textwidth]{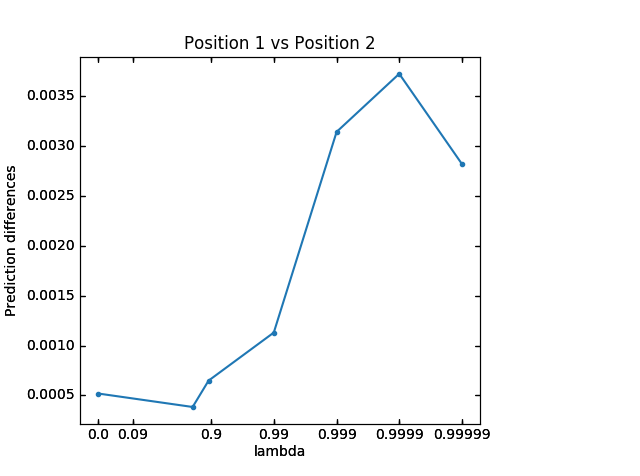}%
\label{fig:synthetic_ranking}}
\caption{Training the ANN model using FL data with two days or 1000 samples.  Both the ANN with bypass (byp) and a variant of the ANN without bypass (\texttt{no\_byp}) results are reported.}
\label{fig:synthetic_main1}
\end{figure*}

To evaluate on an unbiased sample from $\mathcal{D}$, we use the position 1 CTR, 0.464, derived from the last day, $topk_{T-1}$  

 Table \ref{table:dataTable} shows the AUC and Log Loss for the HeldOut data drawn from $\mathcal{D}$ by training a logistic regression model on this dataset.  This is the ideal situation that forms an upper bound on AUC.

 
In addition to the ANN architecture with a Bypass network, we show performance on a variant of the ANN without the Bypass network.  
Figure \ref{fig:synthetic_main1} shows AUCs and MSEs on the FL and RUS datasets at the end of training.  The x-axis is a reverse log scale varying $\lambda$ from 0 to 0.99999.  As $\lambda$ increases, the MSE mostly increases at the expense of FL AUC error.  The ANN with Bypass network's FL MSE error goes back down to 0.00078  (shown in Table \ref{tab:method_MSE}), which is the same performance of a naive method that only averages CTR.  

We note that as $\lambda$ approaches 1, the $Loss_Y$ term in $Loss_N$ becomes diluted, so there should be a set of $\lambda's$ that are optimal in terms of AUC on $\mathcal{D}$, which is seen empirically in Figure  \ref{fig:synthetic_auc_1}.  

The ANN model yields as much as a 12.6\% gain in AUC from the $\lambda=0.9999$ models over $\lambda=0$ on the RUS set and is only 10\% off from a model trained solely on the RUS set.


 
\subsection{Bypass vs non Bypass Results}

The results in Figures \ref{fig:synthetic_main1} show slight improvements using the Bypass vs the non-Bypass ANN both in terms of AUC and higher MSEs on the RUS dataset.  We also analyze the differences in predictions of the bypass network as given different position CTRs.  We feed Position 1 CTR on $day_{T-1}$ as input to the Bypass network along with features to produce predictions, $\hat{\mathbf{y_1}}$ and  do the same for Position 2 CTR on $day_{T-1}$ to create $\hat{\mathbf{y_2}}$.

 We compute the average prediction differences over all trials or $|\hat{\mathbf{y_1}} - \hat{\mathbf{y_2}}|$ for each $\lambda$ value. Figure \ref{fig:synthetic_ranking} illustrates these results.  As MSE increases, so do the difference in predictions.  Therefore, we hypothesize that the Bypass network is increasingly explaining the position CTR in the ANN representation. The non-bypass network, on the other hand, can only produces the same CTR estimate despite different position CTRs. 




\section{Synthetic Evaluations on User Level Bias}
Another factor that causes Position bias may be a User level bias.  Users may be biased towards not clicking on ads below Position 1 regardless of relevance and user interest.  We simulate an additional User level Bias towards Position information by perturbing the Position 2 ranked ads' labels in the previous synthetic evaluations.  Lines 12-14 of Algorithm \ref{alg:synthetic} accomplish this by switching the observed Click label of Position 2 ads from 1 to 0 with probability $r$.  The previous synthetic data generation process in section \ref{sec:synthetic1} is just a special case of this one with $r=0$.  We perform the same experiments as in section \ref{sec:synthetic_main_results} except with a new FL dataset based on r=25\%.  

\subsection{Results}

\begin{figure*}[]
\captionsetup[sub]{font=small,labelfont={bf,sf}}
\centering
\subfloat[\footnotesize{AUC on the FL training set }]{\includegraphics[width=0.25\textwidth]{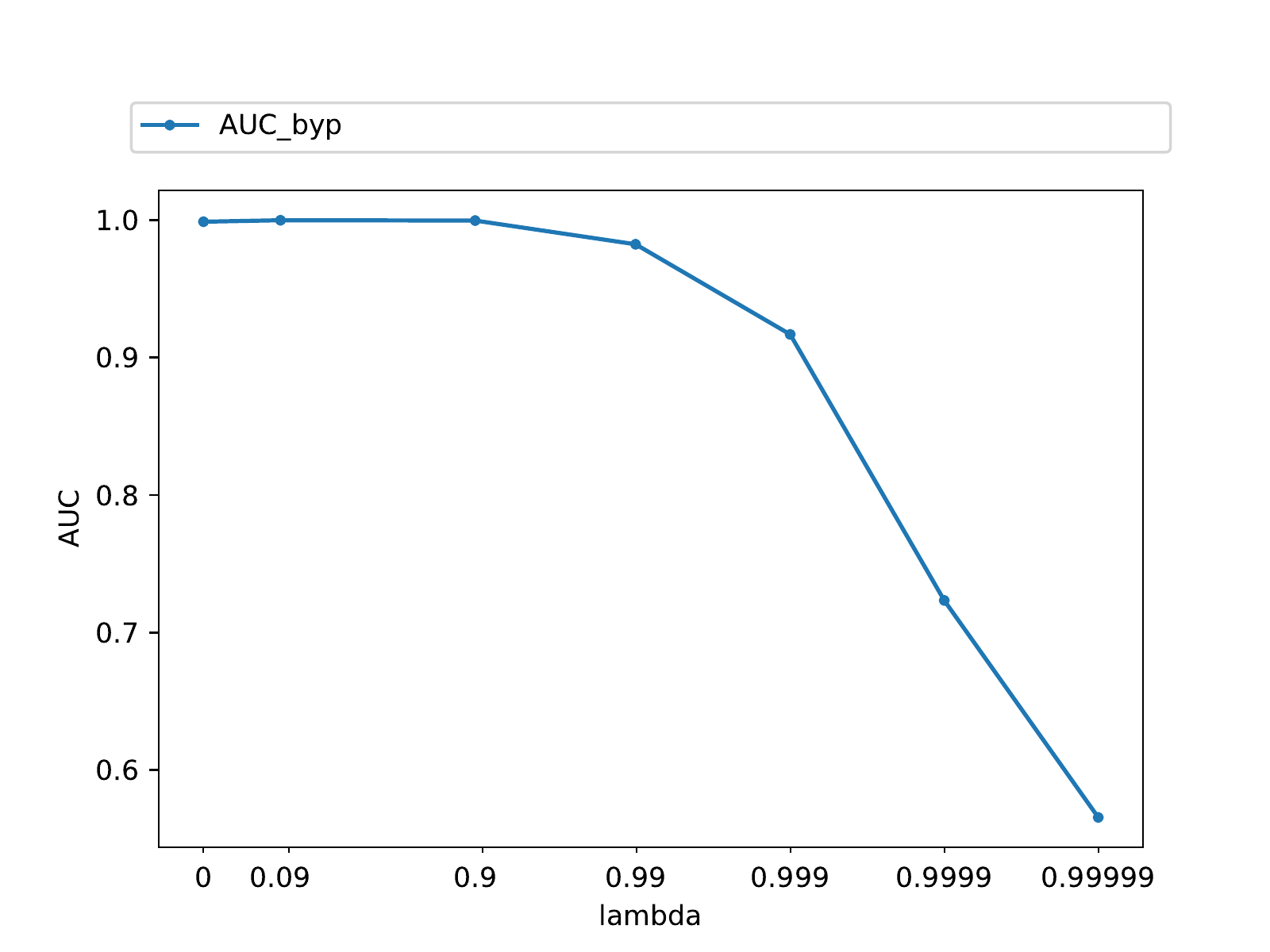}%
\label{fig:synthetic_auc_user_train}}
\subfloat[\footnotesize{MSE on the FL training set  }]{\includegraphics[width=0.25\textwidth]{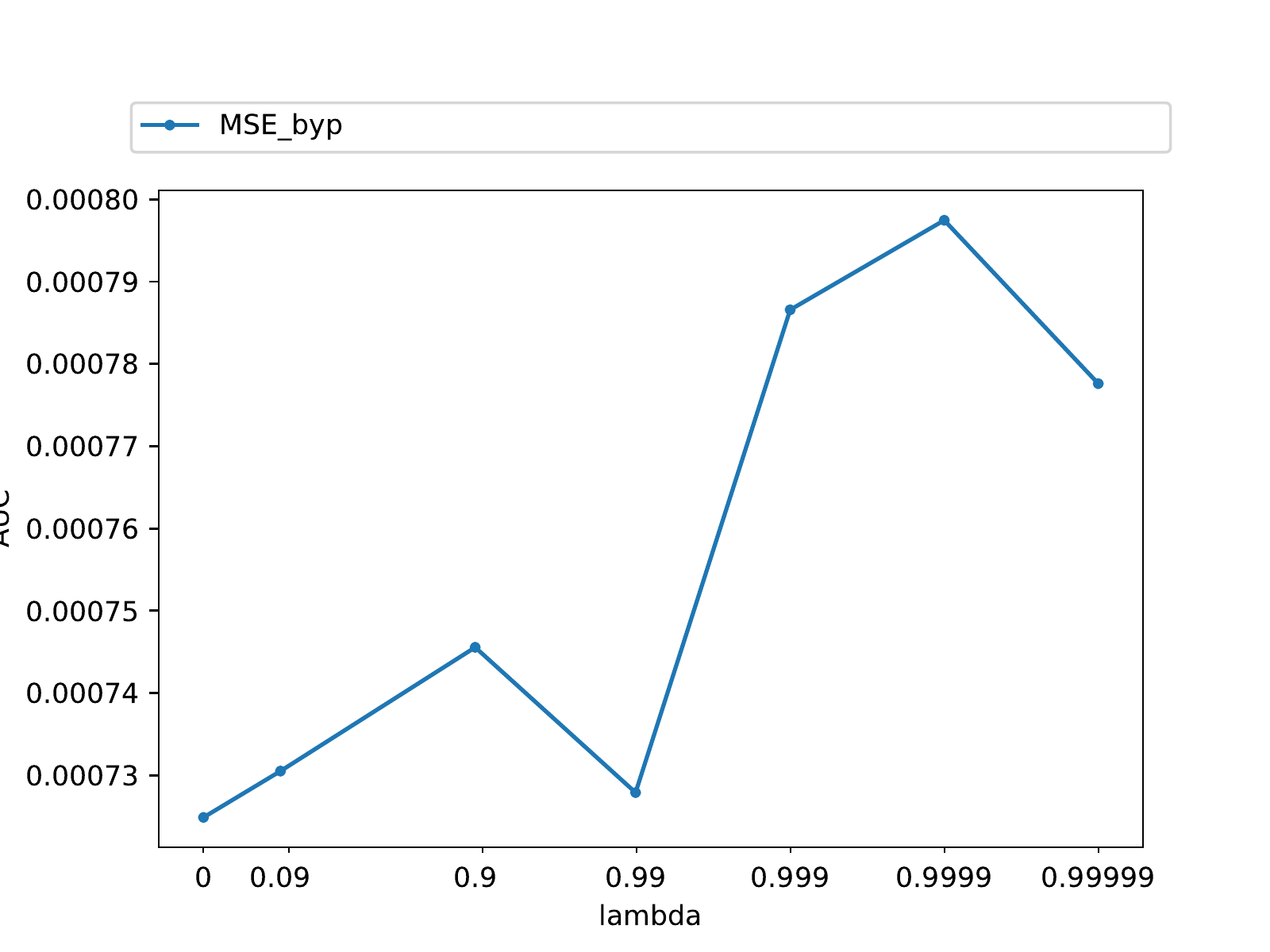}%
\label{fig:synthetic_mse_user_train}}
\subfloat[\footnotesize{AUC on the RUS testing set }]{\includegraphics[width=0.25\textwidth]{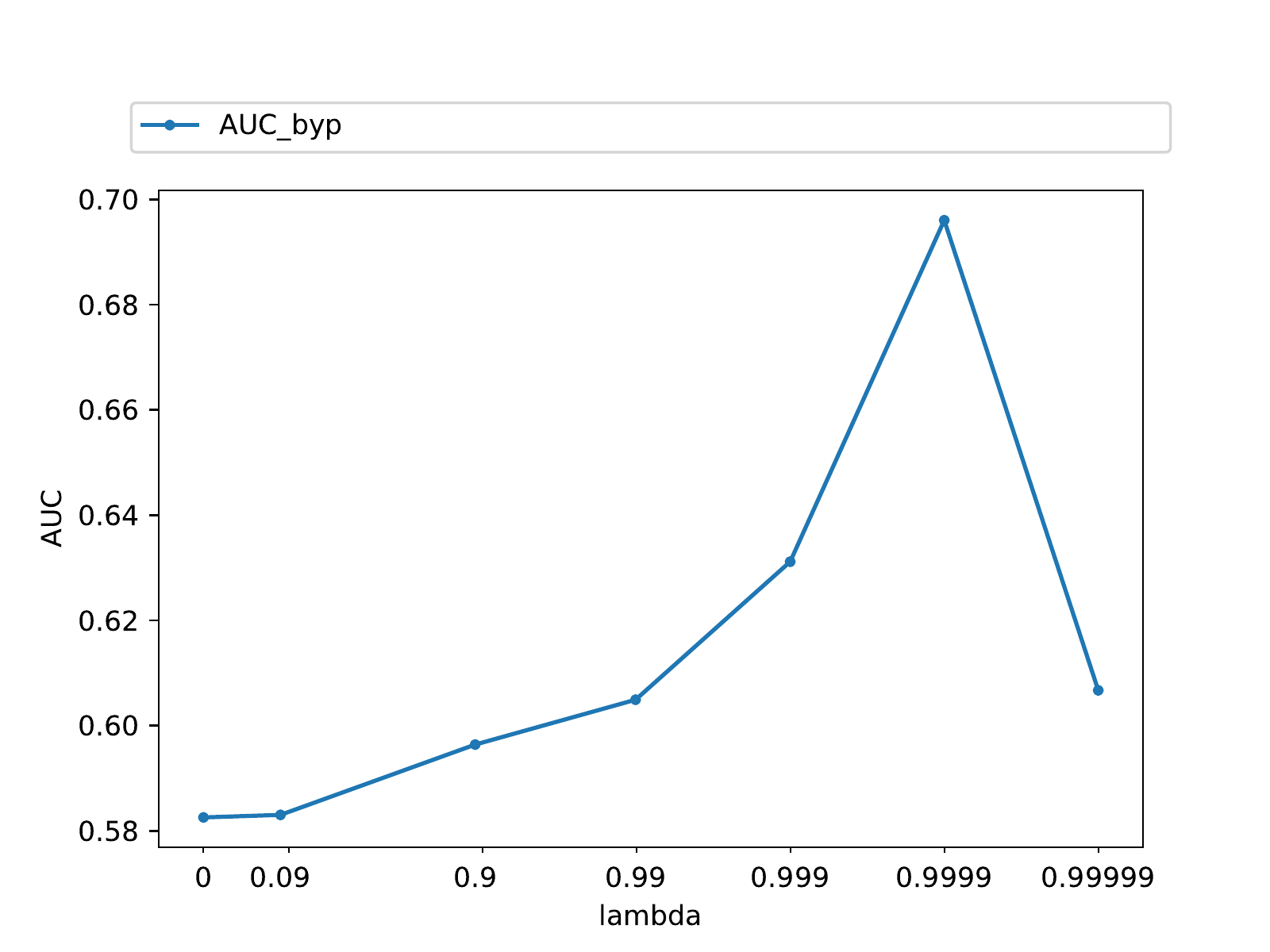}%
\label{fig:synthetic_auc_user_test}}
\subfloat[\footnotesize{MSE on the RUS testing set  }]{\includegraphics[width=0.25\textwidth]{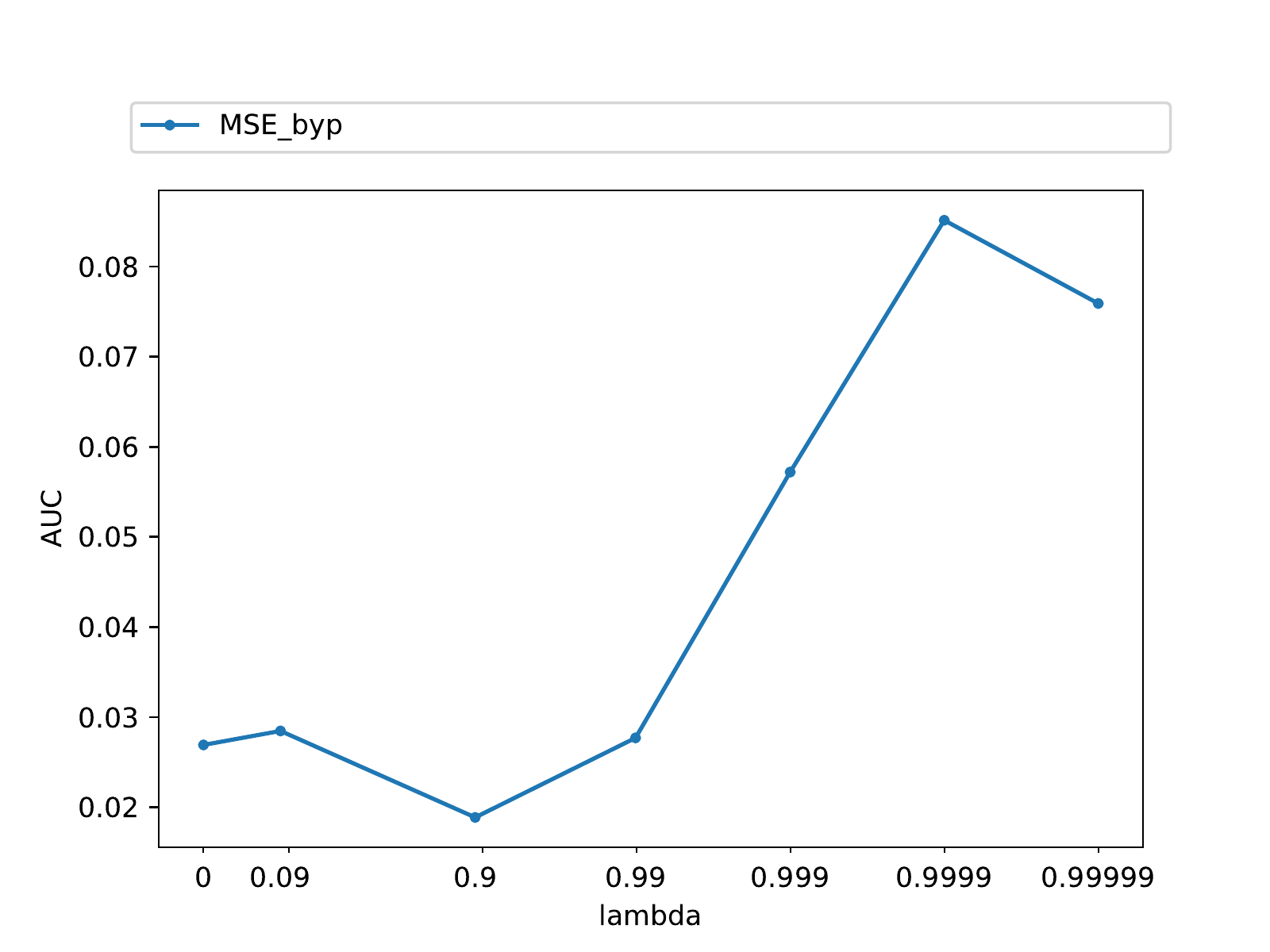}%
\label{fig:synthetic_mse_user_test}}
\caption{Training the ANN model using FL data with r=25\% to simulate a user level position bias. 
 We test on FL and RUS datasets using the vanilla ANN with bypass. }
\label{fig:synthetic_main2}
\end{figure*}

AUCs and MSEs on the FL and RUS data are reported in Figure \ref{fig:synthetic_main2} for both FL and RUS Heldout datasets.  These have similar results to Figure \ref{fig:synthetic_main1}.  Though the highest AUC on the RUS data for r=25\% is less than the highest AUC for r=0\%, there is as much as a  19\% increase between the $\lambda=0$ model's AUC vs the $\lambda=0.9999$ model's AUC.   Therefore, these results empirically indicate that when more bias is added to the data, the ANN representation with appropriate $\lambda$ has higher gains compared to a $\lambda=0$ ANN.

\section{Real World Data Evaluation}
\label{sec:realworld}


In real-world machine learned systems, significant loss in terms of AUC on FL datasets is not desired as samples from this space are often shown online. However, a model, $M_A$, that shows large gains on $\mathcal{D}$ for slight losses on FL data is preferred over a model, $M_B$, that does not show gains over $\mathcal{D}$ since $M_A$ is more likely to perform well online.

In these evaluations, we show AUC gains on $\mathcal{D}$ for acceptable AUC losses on FL data by using datasets from a major search engine's online Ads stack.  The first form of data is an FL dataset consisting of 500 million samples.  The second is an RUS dataset with 100k samples. 


\subsection{Setup}
The hyperbolic tangent function is used for all of the hidden activations except the last layers.  The Base Network is composed of 2 layers with 300 and 150 nodes, respectively.  The Prediction Network has one hidden layer with 300 nodes, while the Bias Network is defined similarly.  The output activation of the Prediction Network is a sigmoid whereas the output activation of the Bias Network is linear.  
We train for 15 epochs on minibatches of size 3072 and evaluate on our RUS dataset. 


\subsection{Main evaluations}

\begin{figure}[t]
    \centering
    \includegraphics[width=0.4\textwidth]{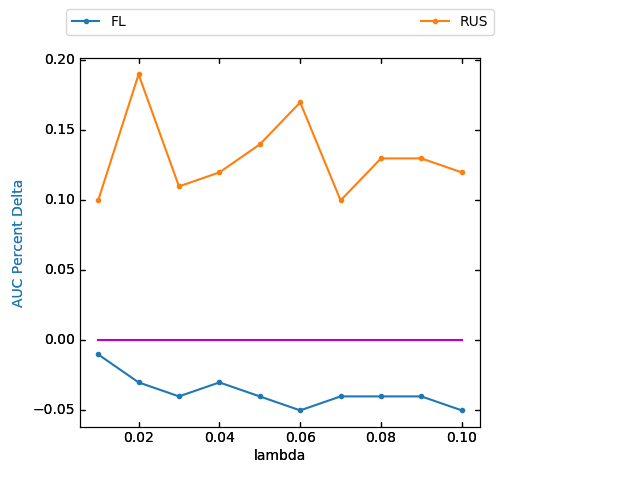}
    \caption{ Absolute percentage difference from the $\lambda=0$ model training on the FL data and testing on FL and RUS}
    \label{fig:main_result}
\end{figure}

We train on our FL dataset for 15 epochs, then test on the RUS and FL datasets as illustrated in Figure \ref{fig:main_result}.  We try varying levels of $\lambda$'s and compare performance to the model with $\lambda=0$.  $b$, which represents position CTR is used as input into the position Bypass Network.

Figure \ref{fig:main_result} shows the AUC percent differences between each model with $\lambda$ value and the $\lambda=0$ model.
This figure shows results for a region where $\lambda$ produces high gains on RUS data, while keeping the error on the FL dataset low.  The FL losses are acceptable in the Ads domain for higher RUS gains.  We see as much as a 0.19 gain with low cost to FL (-0.03) over the $\lambda=0$ model on the RUS data at $\lambda=0.02$.  
These results indicate that the ANN model is generalizing better to  $\mathcal{D}$.

\section{Conclusion}

In this work, we described an Adversarial Neural Network architecture that creates a PClick representation of data with controllable levels of invariance to confounding features.  To show the efficacy of the ANN, we demonstrated evaluations on synthetic and real-world data consisting of as much as 500 million training samples. We believe to the best of our knowledge that the ANN model is the first of its kind to explicitly remove and model feedback loop bias simultaneously.

 To do so, we define an adversarial Bias network that attempts to predict the confounding term, while the Base, Prediction, and Bypass networks attempt to model $y$.  A differentiable squared covariance loss function is used by the Prediction, Bypass, and Base networks to interfere with predictions from the Bias network.  The Bypass network is still able to model the confounding feature linearly and separately.
 Our approach is advantageous to other previously proposed methods since it does not require time and revenue to generate online exploration data.  Rather, it can be used at any point in the natural feedback loop present in an online Ads stack.

\bibliographystyle{ACM-Reference-Format}
\bibliography{prop} 

\end{document}